\documentclass[times,twocolumn,final,authoryear]{elsarticle}

\usepackage{prletters}
\usepackage{framed,multirow}

\usepackage{amssymb}
\usepackage{latexsym}

\usepackage{url}
\usepackage{xcolor}

\definecolor{newcolor}{rgb}{.8,.349,.1}

\journal{Pattern Recognition Letters}

\begin{document}

\ifpreprint
  \setcounter{page}{1}
\else
  \setcounter{page}{1}
\fi
\begin{frontmatter}

\title{Depth Information Guided Crowd Counting for Complex Crowd Scenes}

\author[1]{Mingliang  \snm{Xu}}
\author[1]{Zhaoyang \snm{Ge}}
\author[1]{Xiaoheng \snm{Jiang}\corref{cor1}}
\cortext[cor1]{Corresponding author:}
\ead{jiangxiaoheng@zzu.edu.cn}
\author[1]{Gaoge \snm{Cui}}
\author[1]{pei \snm{Lv}}
\author[1]{Bing \snm{Zhou}}
\address[1]{Zhengzhou University, No.100 of Science Avenue, Zhengzhou 450001, P.R.China}
\author[2]{Changsheng \snm{Xu}}
\address[2]{Institute of Automation, Chinese Academy of Sciences, No.95 of Zhongguancun East Road, Beijing 100190, P.R.China}


\begin{abstract}
It is important to monitor and analyze crowd events for the sake of city safety. In an EDOF (extended depth of field) image with a crowded scene, the distribution of people is highly imbalanced. People far away from the camera look much smaller and often occlude each other heavily, while people close to the camera look larger. In such a case, it is difficult to accurately estimate the number of people by using one technique. In this paper, we propose a Depth Information Guided Crowd Counting (DigCrowd) method to deal with crowded EDOF scenes. DigCrowd first uses the depth information of an image to segment the scene into a far-view region and a near-view region. Then Digcrowd maps the far-view region to its crowd density map and uses a detection method to count the people in the near-view region. In addition, we introduce a new crowd dataset that contains 1000 images. Experimental results demonstrate the effectiveness of our DigCrowd method.
\end{abstract}

%
\end{frontmatter}


\section{INTRODUCTION}
\begin{figure*}[htb!]
\begin{centering}
  \centering
  \includegraphics[width=1.\textwidth]{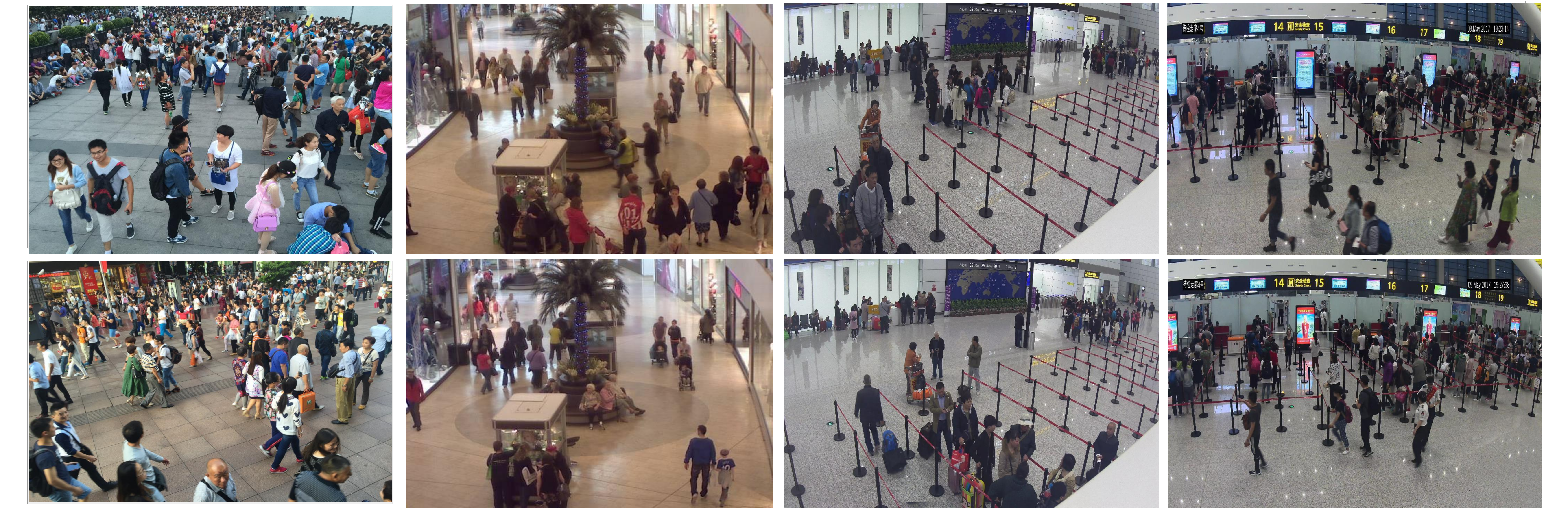}
  \centering
  \caption{The motions of crowds in public areas}
  \centering
  \label{fig:1}
  \end{centering}
\end{figure*}

%
%
%
%
  With the growth of urban populations and the economy,large collections of people in public places is becoming more and more common (Fig.\ref{fig:1}). Therefore, intelligent population analysis systems are of great significance for the establishment of smart cities and safe cities. This is a big challenge to the public security management, which makes the analysis, monitoring, and detection of crowd activity one of the most popular research topics in the intelligent video surveillance field. However, crowd analysis comes with many challenges such as occlusions, high clutter, non-uniform distribution of people, non-uniform illumination, intra-scene and inter-scene variations in appearance, scale and perspective making the problem extremely difficult. In some extended depth of field, we observe that crowd movement lead to an increased number of activities such as aggregation behavior, crowd queuing behavior and the smaller size of the crowd block in far-view region, etc. This crowd activity poses many difficulties for researchers. We define these scenes as EDOF scenes. All of these factors make existing crowd counting systems inadequate.

\begin{figure*}[htb!]
\begin{centering}
\includegraphics[width=1\textwidth]{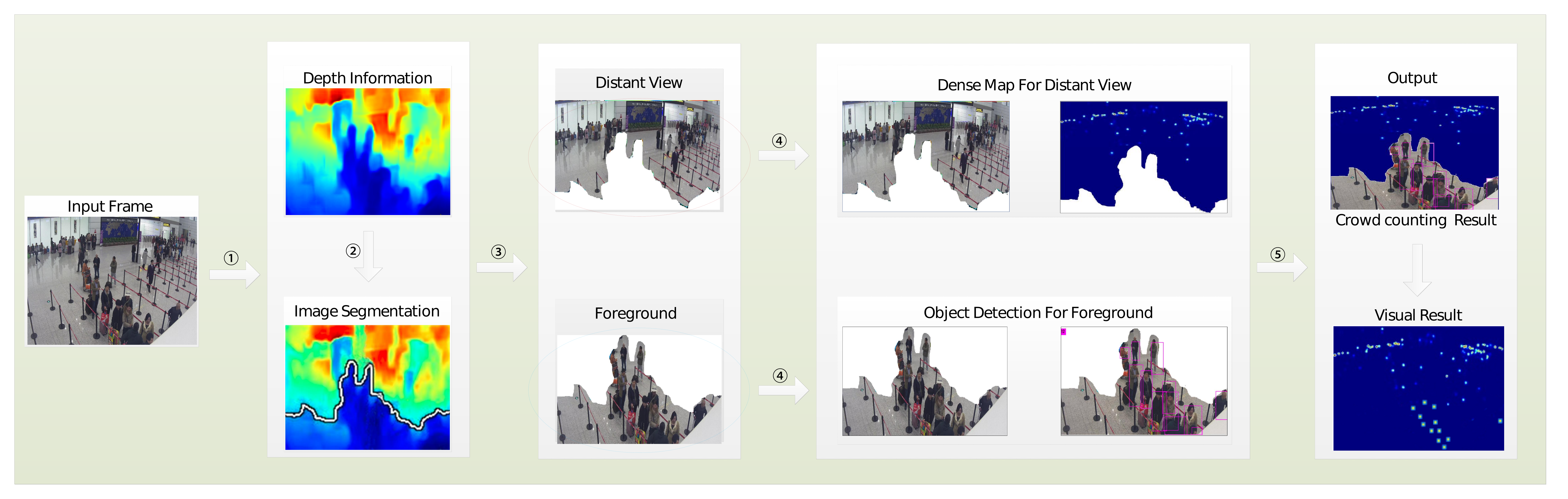}
\caption{The overview of crowd counting for complex crowd scene, which consists of 5 steps: (1) Get image depth information. (2) Two regions are obtained based on pixels with similar local features. (3) Segment input image two parts by depth information. (4) Distant view counting by dense map and foreground counting by object detection. (5) Output image. }
\label{fig:2}
\end{centering}
\end{figure*}

Crowd analysis and its related applications have been deeply studied by researchers in both industry and academia. Based on the limitations of methods or data sources, many researchers focus on a certain type of scenario detection, but these methods do not apply to complex crowd scenarios. The most extensively used methods for crowd counting is regression-based \citep{Redmon2015You,Chan2009Bayesian}. This method is suitable for more crowded or homogeneous conditions, but not for a single individual object in the scene. A method based on a multi-row convolution neural network model is proposed by Shanghai University of Science and Technology~\citep{Zhang2016Single}.  By adding a geometric adaptive Gaussian kernel density map, the corresponding population is calculated by generating the density map method. While the study above provides a good way to detect the number of people in a crowd, the approach is not robust enough when faced with complex scenarios. In order to overcome the problems above, a new mixed method for crowd counting in complex crowd scenes is proposed in this paper. Fig.\ref{fig:2} illustrates the overview of the proposed algorithm. Specifically, first of all, a depth map is used to divide the complex crowd scene into a near-view region and a far-view region. The depth map algorithm gets the depth information of the image, and the purpose is to divide the image into a near-view region and a far-view region fixed.  Image segmentation based on the yolo framework method is used to detect the effect of the picture, and not every scene is quantitatively divided. We propose two reasons for the failure of yolo methods. One is non-even distribution of people and non-even pixels of person, the other is the existence of aggregation of the crowd, and the screening effect of serious leak detection. These reasons lead to poor results. Then, a density map method is used for the crowd statistics of the far-view region of the image. Based on the mixed method, the results are automatically merged, and are closer to the actual situation.

Compared with previous work, the contributions of this paper lie in the following aspects:

1) For the representation of crowds in public areas, a concept of complex crowd scenes is proposed, and a crowd density detection system based on depth neural network for video streaming mixed method is proposed.

2) A kind of image segmentation method based on image depth information is proposed, and the image is divided into two parts (near-view region and far-view region) as the input for the neural network. And the spatial context method is used to eliminate duplicate detection.

3) We provide a dataset of complex crowd scenes from an airport and provide a stable, reliable and easily applied method for crowd counting in public places.

The remainder of this paper is organized as followed: Section 2 introduces some background and related work on crowd counting for dense map and crowd analysis. In Section 3, we describe the number of images based on depth information segmentation, the spatial context method, object detection, the density map method, and the total number of images. In Section 4, the experiment results are analyzed. We evaluate the performance of the proposed algorithm.


\section{RELATED WORK}
Crowd counting has been tackled in computer vision by a myriad of techniques. At present, the crowd counting methods fall into the following categories: detection-based approaches, regression-based approaches, density estimation-based approaches and CNN-based approaches\citep{Sindagi2017A}.

Most research focuses on a detection style framework, where a sliding window detector is used to detect people in the scene \citep{Wojek2012Pedestrian} and this information is used to count the number of people. Another method of overall detection of the image uses features extracted from a full body \citep{Dalal2005Histograms,Leibe2005Pedestrian,Tuzel2008Pedestrian,Enzweiler2009Monocular}, and then uses the learning method \citep{Yang2015Multi,Chang2017Semisupervised} to count the number of people. Though successful in low density crowd scenes, these methods are adversely affected by the presence of high density crowds. To overcome these issues, researchers attempted to count by regression where they learn  mapping \citep{Chang2016Compound} between features extracted from local image patches to their counts, such as foreground, edge, texture, gradient and other characteristics, and remove the background interference, and try to count  large areas of high-density population\citep{Chan2009Bayesian,Ryan2010Crowd,Chan2008Privacy,Shao2016Slicing,Chen2012Feature}. No detection method is reliable enough to provide sufficient information for accurate counting of high density crowds due to various reasons, such as low resolution, severe occlusion, foreshortening and perspective. Idrees et al.\citep{Idrees2013Multi} observed that there exists a spatial relationship that can be used to constrain the count estimates in neighboring local regions. With these observations in mind, they proposed to extract features using different methods that capture different information. By treating densely packed crowds of individuals as irregular and non-homogeneous texture, they employed Fourier analysis along with head detections and SIFT interest point-based counting in local neighborhoods.The method is well applied in a wide range of high density populations.

While the earlier methods were successful in addressing the issues of occlusion and clutter, most of them ignored important spatial information as they were regressing on the global count. Lempitsky et al.~\citep{Lempitsky2010Learning} proposed  a linear mapping between local patch features and corresponding object density maps, thereby incorporating spatial information into the learning process. And Phamet al~\citep{Pham2015COUNT} tackled the problem of large variation in appearance and shape between crowded image patches and non-crowded ones by proposing a crowdedness prior and they trained two different forests corresponding to this prior. The above method in any region of a density map can calculate the number of objects in the region and optimize the problem of counting the population.

Researchers use the CNN \citep{Chang2017Semantic} to connect the nonlinear function from the crowd image to the corresponding density map or the corresponding counting ability. It is through the depth of learning network to extract the image to be detected feature map (energy map, density map) to be integrated, so as to make the number of estimates. However, according to the limitations of the convolution kernel, the work can only determine the effect of a single scene to deal with, such as the crowd is extremely dense, or the crowd sparse, there is no aggregation phenomenon, so we have to solve the problem is invalid. Zhang et al.~\citep{Zhang2015Cross} analyzed existing methods and identified that their performance reduces drastically when applied to a new scene that is different from the training dataset. Boominathan et al.~\citep{Boominathan2016CrowdNet} combined deep and shallow fully convolutional networks to predict the density map for a given crowd image. The combination of two networks enables one to build a model robust to non-uniform scaling of crowd and variations in perspective. To overcome this issue, they proposed mapping from images to crowd counts and to adapt this mapping to new target scenes for cross-scene counting.

These methods were a good solution for crowd counting in some difficult situations. However, we found that when the dataset contains a distant view of the scene, and more complex types of population density occurs, often the above cannot deal with the crowd counting problem.

\section{OUR PROPOSED SYSTEM}
Through the analysis of specific scenes, we found that airports areas are complex and different form other existing dataset. There are many forms of crowd movement in EDOF scenes, and only using one method of crowd counting cannot achieve accuracy. We propose a new fusion method system of the specific scenes. Specifically:

 1) We use a new segmentation method, first obtaining the depth information of the input image based on the neural network, and then using the local similarity of the deeper information to classify the image into the near-view region and the far-view region.

 2) For the near-view region, our method uses deep convolutional neural networks detect people. Due to its fast network operation, advanced object detection is still maintained. We also apply spatial constraints to solve the problem of repetitiveness detection at the edges of image segmentation.

 3) The Yolo detection method cannot deal with distant conditions. Because of the far-view region, there will be a serious occlusion or aggregation phenomenon. So we use the dense map method which fits to calculate small distant targets. This gives a better experimental effect.


\subsection{Image segmentation}

For our framework, we need to find the lines to accurately divide the image, and accurately assign them to the two different methods of detection. We use the segmentation method of the neural network to get the depth map of the scene~\citep{Zbontar2015Computing, Lecun2016Stereo}.

1) Firstly, we predict depth map from a single image using a multi-scale deep network. We use the work from\citep{Eigen2014Predicting}. The model is a multi-scale deep network that first predicts a coarse global output based on the entire image area, then refines it using finer-scale local networks. The model improved from \citep{Eigen2014Depth}. They make the model deeper (more convolutional layers)  and try best to output high resolution. Third, instead of passing output predictions from scale 1 to scale 2, we pass multichannel feature maps; because they could train the first two scales of the network jointly from the start, simplifying the training procedure and yielding performance gains.

\begin{figure}[htb]
 \begin{centering}
  \centering
  \includegraphics[width=0.49\textwidth]{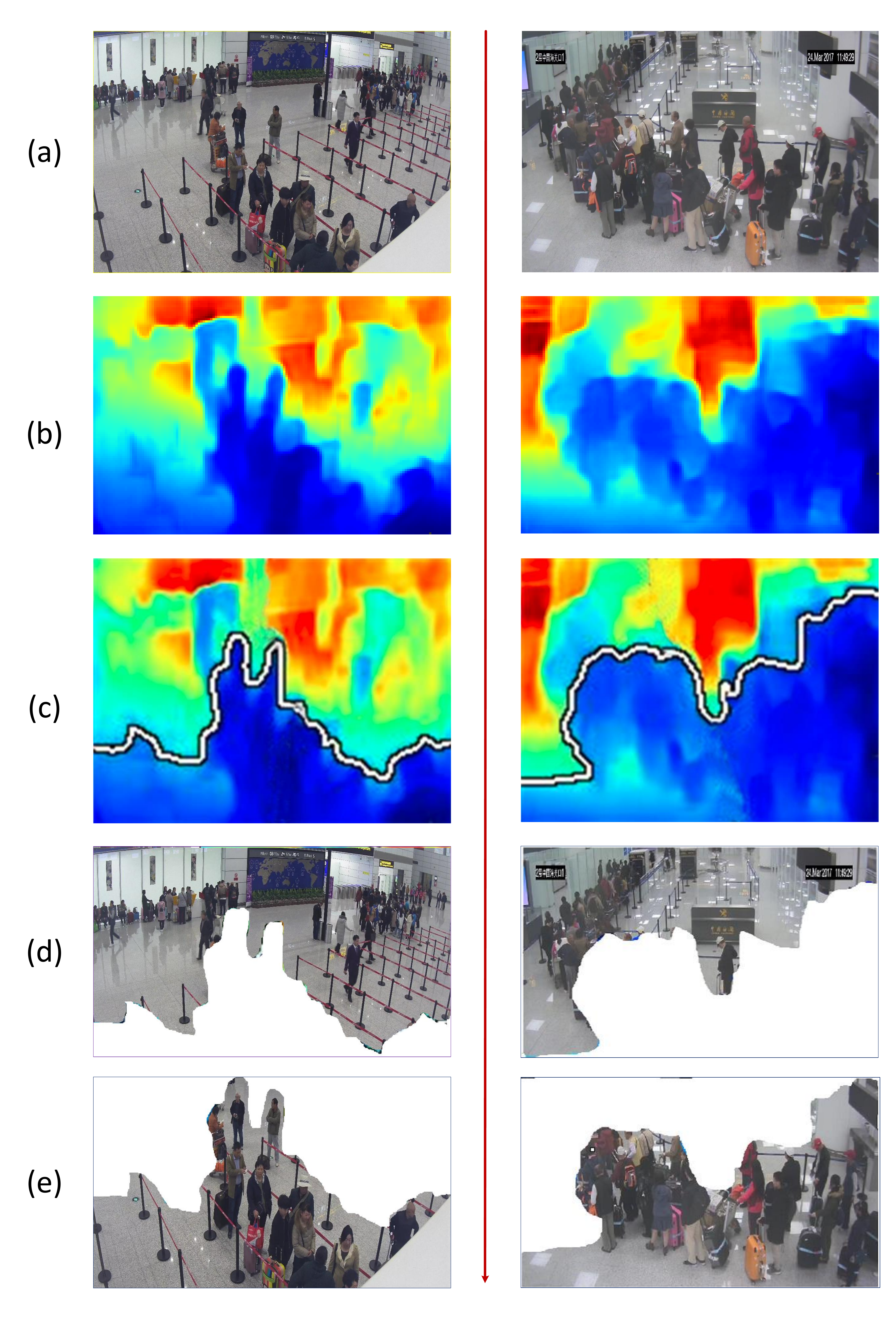}
  \centering
  \caption{Segmentation based on depth information.Using local similarity of depth information divide into two parts.((a)Input image (b) Depth information (c) Split line (d) Far-view region (e) Near-distance part  )  }
  \label{fig:3}
  \end{centering}
\end{figure}

2) Second, we use an approach to generate the near-view area and the far-view area based on their color similarity and proximity in the image. This is done in the five-dimensional $[labxy]$ space, where $[lab]$ is the pixel color vector in the color space, which is widely considered as perceptually uniform for small color distances, and x, y is the pixel position. While the maximum possible distance between two colors in the color space is limited, the spatial distance in the x, y plane depends on the image size. We use a distance measure that considers superpixel size. Using it, methods enforce color similarity as well as pixel proximity in the 5D space such that the expected patch sizes and their spatial extent are approximately equal~\citep{Achanta2010SLIC}.
In order to ensure the integrity of the picture information, we divide the image for the two parts by the rule. Fig.\ref{fig:3} shows our segmentation rule to our scenes.

\subsection{Detection on near-view region}
\begin{figure}[htb]
 \begin{centering}
  \centering
  \includegraphics[width=0.49\textwidth]{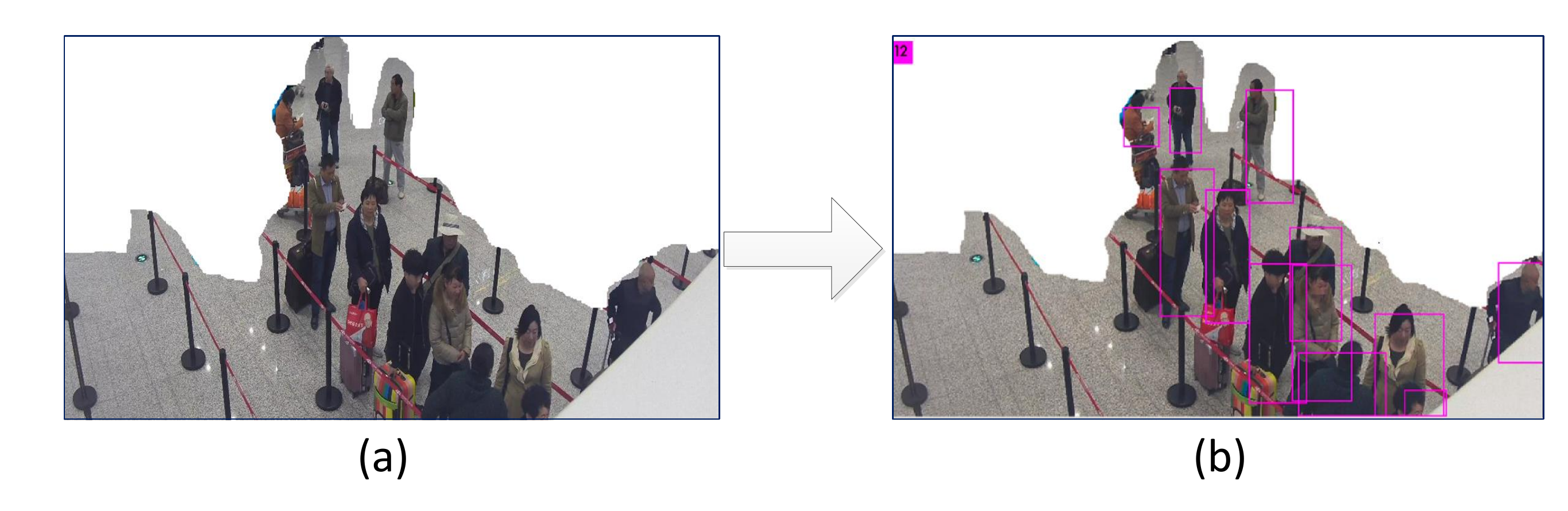}
  \centering
  \caption{Object detection for the foreground result ((a) Input foreground image by depth information segmentation; (b) Object detection rusult) }
  \label{fig:4}
 \end{centering}
\end{figure}

We first pre-train weights with a traditional CNN for general feature learning. The convolutional neural network takes a video frame as its input and produces a feature map of the image. Once the pre-trained weights are able to generate visual features, we adopt the YOLO architecture as the detection module~\citep{Redmon2015You}. On top of the convolutional layers, YOLO adopts fully connected layers to regress feature representation into region predictions. It denotes that the image is divided into {S*S} splits. Each split has B bounding boxes predicted, represented by its 5 location parameters, including x, y , w , h , and its confidence c. A one-hot feature vector of length C is also predicted, indicating the class label of each bounding box. In our framework, we follow the YOLO architecture and set S=7, we only focus on one person in our image, so we set the class to one. At test time we multiply the conditional class probabilities and the individual box confidence predictions:

\begin{small}
\begin{equation}
\resizebox{.9\hsize}{!}{$Pr(Class{s_i}|Object) \times Pr(Object) \times IOU_{pred}^{truth}=Pr (clas{s_i}) \times IOU_{pred}^{truth}$}
\end{equation}
\end{small}
which gives us class-specific confidence scores for each box. These scores encode both the probability of that class appearing in the box and how well the predicted box fits the object. Fig.\ref{fig:4} is our result of near-view region detection.
\subsubsection{spatial constraining}

By analyzing the experimental data, we find that there are often repeated detection problems near the dividing line. As shown in Fig.5(a). Through analyzing the surveillance video and the statistics of the false detection area for a long time, we find the range of false detection area in the surveillance video is relatively fixed. For example, under normal circumstances, the cutting line will cut through the individual, and repeat count in the near-view region count and the far-view region count. Therefore, we try to process the original test results in the image to eliminate some false detections.

Scene understanding and segmentation is a popular area of research in computer vision. Many related works, Zia et al.~\citep{Zia2015Towards} and Brun et al.~\citep{Brun2014Dynamic} have been proposed, but there are still some difficulties when scene understanding and segmentation is applied to EDOF scenes. First, it will cause fatal error detection once the scene understanding and segmentation for EDOF scenes is incorrect. Second, the correct region can't be computed easily according to the state of the crowd (occluded, imbalanced or small). On the contrary, the region where the crowd appears may be judged as an invalid area, which will increase the probability of missing detection. At present, there is still a lack of an effective algorithm to segment the EDOF scenes accurately and automatically. In order to handle those uncorrected cases, we manually segment the EDOF scenes. Fig.5 shows the flow of imposing spatial constraints, the specific steps are as follows:

Step 1: Input the first frame of surveillance video for each scene containing crowd to be detected.

Step 2: Draw a continuous split polyline manually and define a region where there may be a false detection, taking the priori information of the crowd space distribution into account. The polyline equation is:

\begin{equation}
\left\{ {\begin{array}{*{20}{c}}
{{y_1} = {k_1}x + {b_1},x \in [{x_0},{x_1});}\\
{{y_2} = {k_2}x + {b_2},x \in [{x_1},{x_2});}\\
{...}\\
{{y_n} = {k_n}x + {b_n},x \in [{x_{n - 1}},{x_n}].}
\end{array}} \right.
\end{equation}

The polyline equation will vary with different scenes.

Step 3: Introduce a multi-clue detection scheme to postprocess the detection results. Assume the coordinates of the upper left corner and the lower right corner of the bounding box of detection results are $({x_{\min }},{y_{\min }})$ ,$({x_{\max }},{y_{\max }})$, then the center coordinates of the bounding box are ${x_c}{\rm{ = }}({x_{\min }} + {x_{\max }})/2$ ,${y_c}{\rm{ = }}({y_{\min }} + {y_{\max }})/2$. If ${x_c} \in [{x_{i - 1}},{x_i})$, and ${y_c} < {y_i}$ ,$i \in (1,n]$, it means the bounding box appears in invalid region. This result will be deleted directly.

Step 4: Directly deleted the bounding box appeared in invalid region. Visualize the final detection results (Fig.5(b)).

\begin{figure}[htb]
\begin{centering}
  \centering
  \includegraphics[width=0.49\textwidth]{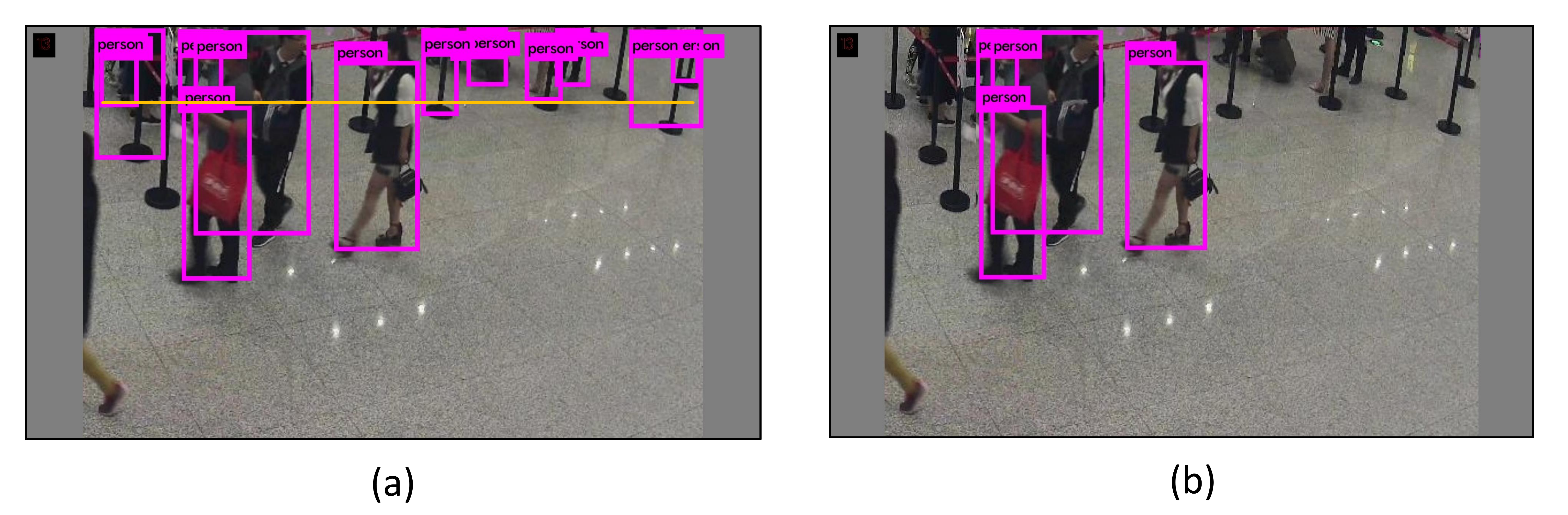}
  \centering
  \caption{The flow of imposing spatial constraint. (a) is the detection without patial constraint. (b) is the detection result after imposing spatial constraint on image (a). }
  \label{fig:5}
  \end{centering}
\end{figure}
\subsection{Density map estimation on far-view region}

The Yolo module is not good for far-view region which each other's objects and very small aggregation phenomenon because of the grid in the prediction box. So for the far-view region of the crowd, we use the dense map in our system for detection. The system uses the convolution neural network, the output image is a density map of the crowd, and then the real count is obtained by integration.

We describe how to convert an image with people's heads to a map of crowd density. At the same time we make the density graph calculation result better match the number of people in the real picture. To convert this to a continuous density function, first we denote the distance to its $k$ nearest neighbors as $ { d_1^i,d_2^i,...,d_m^i}$ , The average distance is therefore
\begin{equation}
\overline {{d^i}} {\rm{ = }}\frac{1}{m}\sum\nolimits_{j = 1}^m {d_j^i}
\end{equation}shanghaitech may convolve this function with a Gaussian kernel ${G_\sigma }$, so that the density is
\begin{equation}
F(x) = \sum\limits_{i = 1}^N {\delta (x - {x_i})*{G_{\sigma i}}(x),} {\sigma _i} = \beta \overline {{d^i}}
\end{equation}
When there is a person at pixel ${x_i}$ , we represent it as a delta function $\delta (x - {x_i})$, $N$  is the total number of people in the image for some parameter β. In other words, we convolve the labels H with density kernels adaptive to the local geometry around each data point, referred to as geometry-adaptive kernels. In our experiment, we have found empirically $\beta= 0.3\ $ gives the best result. In Fig. \ref{fig:5}, we have shown density maps of one image in our dataset.

\begin{figure}[htb]
 \begin{centering}
  \centering
  \includegraphics[width=0.49\textwidth]{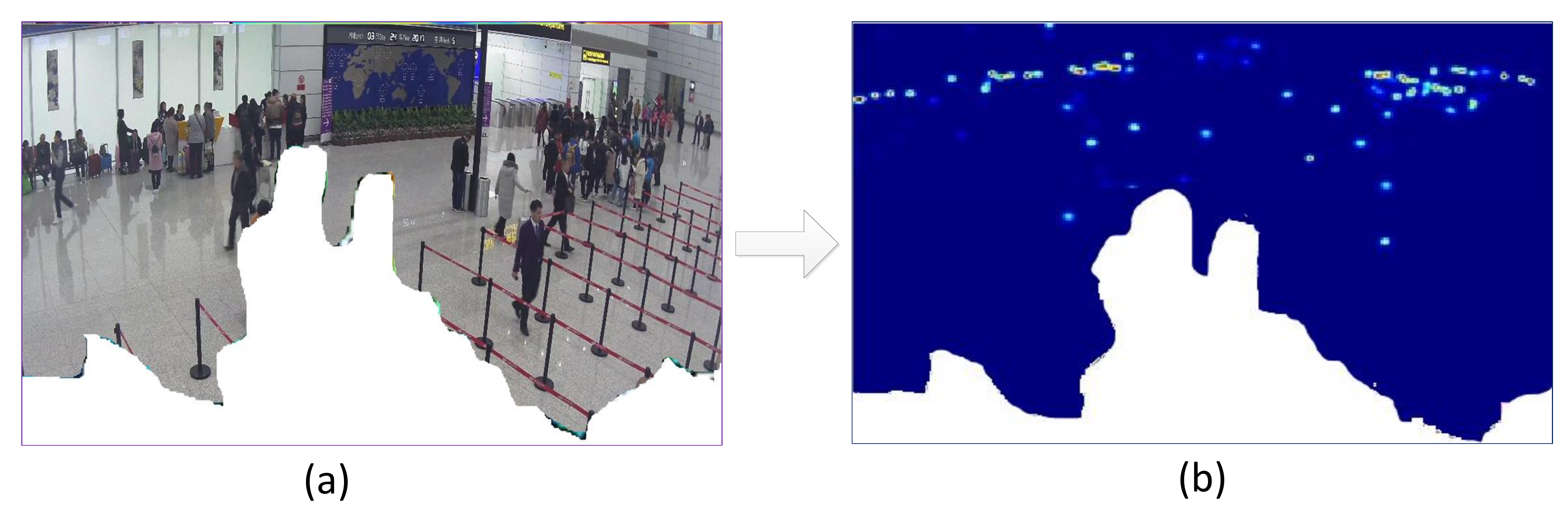}
  \centering
  \caption{ Our methods result for far-view region. ((a) Original far-view region images; (b) Crowd density maps). }
  \label{fig:6}
 \end{centering}
\end{figure}

\subsection{System work results}

The advantage of our method is that there are specific detectors in a particular area. Our system detects images by dividing them into near-view region and far-view region. So, we do not need to constantly modify the parameters to match datasets which contain heads of very different sizes. Experimental results show that our system greatly improves detection accuracy.

%

\begin{figure}[htb]
\begin{centering}
  \centering
  \includegraphics[width=0.49\textwidth]{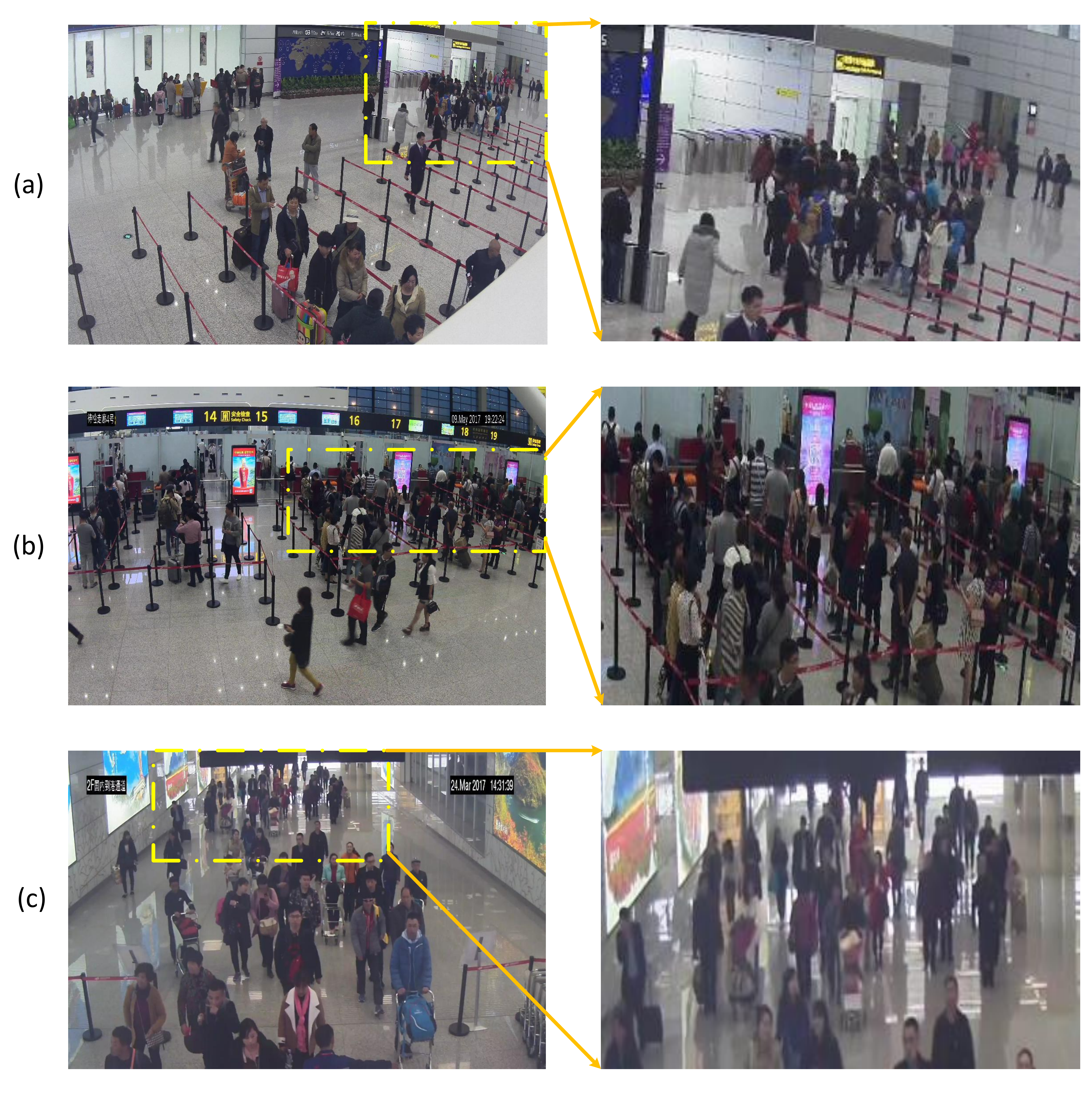}
  \centering
  \caption{The details of our dataset. There are great differences between the near-view region and far-view region. (a) Airport Hall:  In the near-view region, there is random distribution of people. But in the far-view region people gather, with occlusions. (b) Check-in Office: The scene is mainly comprised of queueing people. (c) Corridor: The scene is more distant than the others. There is significant variation in the scale of the people in the images.}
  \label{fig:7}
  \end{centering}
\end{figure}
\section{EXPERIMENTAL RESULT}
We evaluate our mixed methods on three different dataset, two existing dataset and our own dataset. The existing dataset is chosen from Zhang et al.~\citep{Zhang2016Single}, and another existing dataset is the Mall dataset~\citep{Chen2012Feature}. We use the training network on the ciisr scene dataset directly for other dataset. At the end of our experiment, we have better results on these dataset. The experimental environment configuration: Intel Xeon(R) CPU E5-1620 v4@3.50GHz, 64G memory, TITAN X.

\subsection{Evaluation metric}
Following the convention of existing works for crowd counting. We use Mean Absolute Error (MAE) and Mean Squared Error (MSE) as the metric for comparing the performance of our methods against the state-of-the-art crowd counting methods. Roughly speaking, MAE indicates the accuracy of the estimates, and MSE indicates the robustness of the estimates.
both the absolute error:
\begin{equation}
MAE = \frac{1}{N}\sum\limits_1^N {|observed - predicted|}
\end{equation}
and the mean squared error:
\begin{equation}
MSE = \sqrt {\frac{1}{N}\sum\limits_1^N {{{(observed - predicted)}^2}} }
\end{equation}
which are defined as follows: where $N$ is the number of test images, ${observed}$ is the actual number of people in the image, and ${predicted}$ is the estimated number of people in the image.

\subsection{Ciisr dataset}
As existing dataset are not entirely suitable for evaluation of the crowd count task considered in this work, we introduce new EDOF scenes from an airport for a crowd counting dataset named ZZU-CIISR which contains three different scenes: Airport Hall, Check-in Office, Corridor. This dataset contains 20 videos with resolution 1920$\times$1080. Each video random sampling 200 frames and depth map of field 30 to 50 meters. This dataset contains 1000 images, with about of 120 people with EDOF scenes in each image.
\begin{figure*}[htb!]
\begin{centering}
  \centering
  \includegraphics[width=1\textwidth]{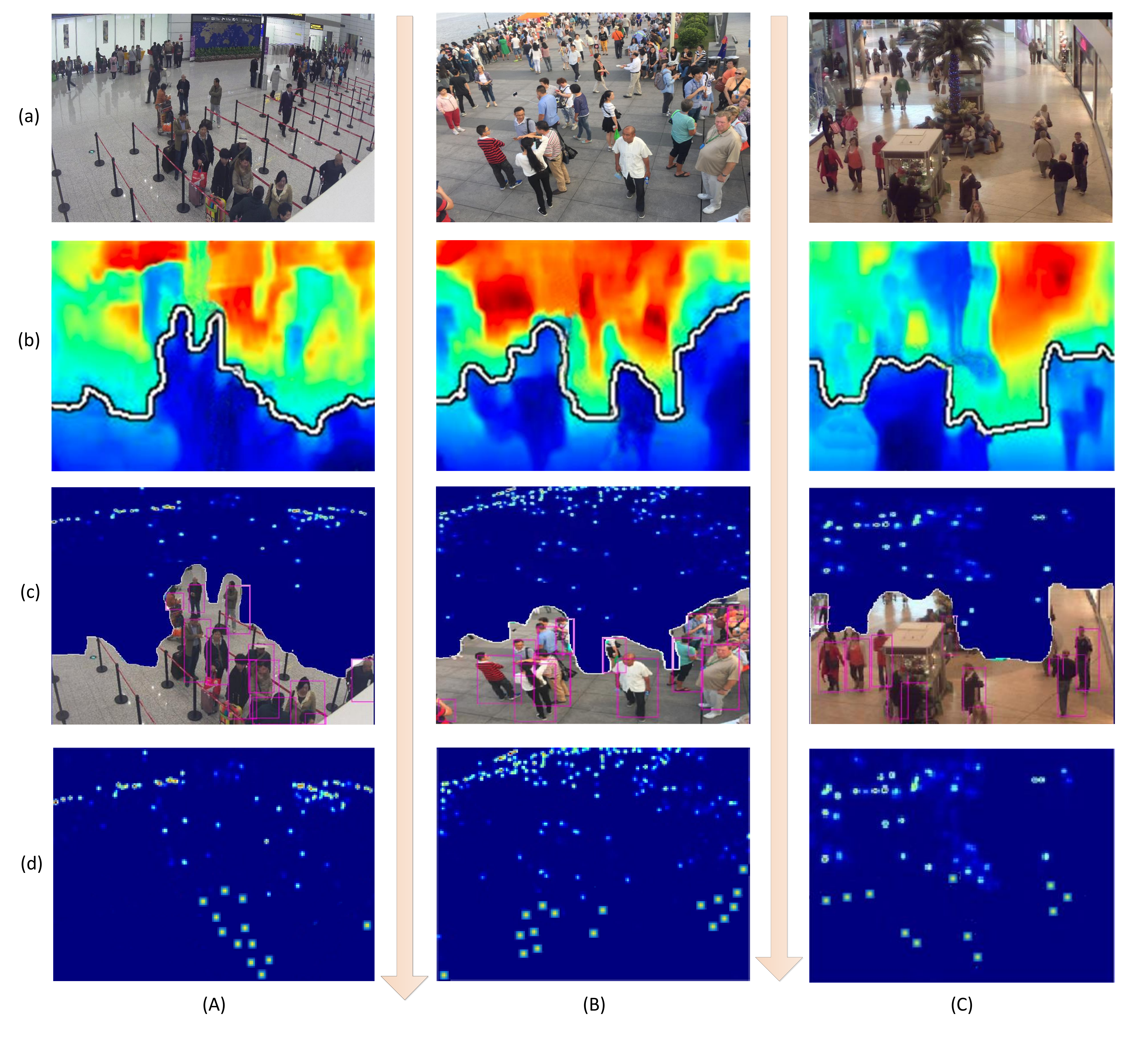}
  \centering
  \caption{Our system in different dataset. Vertical: (A) from our ciisr dataset. (B) from mall dataset. (C) from shanhaitech dataset. Horizontal: (a) The input image. (b) The depth map from input image. (c) Our segmentation rule. (d) Our system result. (e) Visualization of results.}
  \label{fig:8}
  \end{centering}
\end{figure*}

Each scene of the new dataset was captured using a surveillance camera in an airport with challenging lighting conditions and glass surface reflections. There is a clear difference between our dataset and other dataset. The CIISR dataset covers diverse crowd densities, from sparse to crowded, as well as different activity patterns (static and moving crowds) under a larger range of illumination conditions at different times of the day. In addition, in comparison to the other dataset, the CIISR dataset experiences more severe perspective distortion, which causes larger changes in size and appearance of objects at different depths , and has more frequent occlusion problems caused by the scene objects, e.g. in the near-view region, people in our EDOF scenes have a larger population of pixels. In the far-view region, the crowd is very dense, and the head of the crowd occupies a more uniform pixel. The details of our dataset is shown in Fig \ref{fig:7}. Table 1 gives the statistics of CIISR dataset and its comparison with other dataset.
\begin{table}[htbp]
\normalsize
\renewcommand\arraystretch{1.5}
\caption{\upshape Comparation of CIISR dataset with existing dataset: Num is the number of images; Ave is the average crowd count.}

\centering  
\begin{tabular}{|l|c|c|c|}  
\hline
\cline{1-4}
Dataset &Resolution &Num &Ave   \\ \cline{1-4} 

Shanghaitech &768$\times$1024 &716 &123.6  \\   \cline{1-4}      

Mall Dataset &640$\times$480 &2000 &--  \\   \cline{1-4}     

CIISR Dataset &1080$\times$720 &1000 &117   \\  \cline{1-4}

\end{tabular}
\end{table}

We compare our method with other works for our dataset. We first compare it with the work of Zhang et al.~\citep{Zhang2016Single}, which also uses dense map for crowd counting and achieved state-of-the-art accuracy at the time. We also compare our work with ~\citep{Redmon2015You}, which is a new approach to object detection. They frame object detection as a regression problem to spatially separated bounding boxes and associated class probabilities. It has state-of-the-art object detection performance.

The comparison of the performances of all the methods on the dataset is shown in Table 2. There is no doubt that our method achieves the highest accuracy, compared with the other three methods. And our method has the best mae/mse on our dataset. We effectively use the segmentation rules to accurately predict the number of people sheltered and the crowd in the far-view region. The result of this dataset is shown in Fig. \ref{fig:8}(a).

\begin{table}[]
\centering
\caption{\upshape Comparing performances of different methods on CIISR dataset}
\label{my-label}
\begin{tabular}{|l|l|l|l|l|l|l|l|}
\hline
\multicolumn{2}{|l|}{Methods}  & \multicolumn{2}{c|}{Hall}                           & \multicolumn{2}{c|}{Check in office}                & \multicolumn{2}{c|}{Corridor}                       \\ \hline
\multicolumn{2}{|l|}{}         & \multicolumn{1}{c|}{MAE} & \multicolumn{1}{c|}{MSE} & \multicolumn{1}{c|}{MAE} & \multicolumn{1}{c|}{MSE} & \multicolumn{1}{c|}{MAE} & \multicolumn{1}{c|}{MSE} \\ \hline
\multicolumn{2}{|l|}{Yolo}     & 18.42                    & 22.02                    & 30.50                    & 35.70                    & 7.00                     & 7.78                     \\ \hline
\multicolumn{2}{|l|}{Dense}    & 16.70                    & 19.11                    & 11.56                    & 16.10                    & 10.67                    & 12.28                    \\ \hline
\multicolumn{2}{|l|}{DigCrowd} & \textbf{7.57}            & \textbf{9.11}            & \textbf{9.37}            & \textbf{11.05}           & \textbf{3.30}            & \textbf{4.76}            \\ \hline
\end{tabular}
\end{table}

%
%
%
%
%
%
%

%
%
%
%
\subsection{The Shanghaitech dataset}
We also evaluate our method on the Shanghaitech dataset. This dataset was first introduced by Zhang et al.\citep{Zhang2016Single}. This dataset consists of two parts: there are 482 images in Part A which are randomly crawled from the Internet, and 716 images in Part B which are taken from the busy streets of metropolitan areas in Shanghai. The crowd density varies significantly between the two subsets, making accurate estimation of the crowd more challenging than in most existing dataset.
\begin{table}[]
\small
\renewcommand\arraystretch{1.5}
\caption{ \upshape Comparing performances of methods on Shanghaitech dataset.}
\centering  
\begin{tabular}{|p{4.25cm}|p{1cm}|p{1cm}|}  
\hline
\cline{1-3}
scenes &MAE  &MSE  \\ \hline  

Zhang et al.~\citep{Zhang2015Cross} &32.0 &49.8  \\ \cline{1-3}        

MCNN~\citep{Zhang2016Single} &26.4  &41.3  \\ \cline{1-3}       

YOLO~\citep{Redmon2015You} &83.9  &111.3 \\ \cline{1-3}

DigCrowd &\textbf{21.5} &\textbf{32.3}\\  \hline
\end{tabular}
\end{table}
We compare our method with three existing methods on the shanghaitech $ part_B $ dataset, which satisfies the assumptions of EDOF scenes. Table 3 shows the performance of our method and other methods on this dataset. Yolo~\citep{Redmon2015You} detection of the foreground is better but not good for whole image. MCNN~\citep{Zhang2016Single} use cnn to get density map estimation to obtain better head detection results in crowd scenes. The work of ~\citep{Zhang2015Cross} is based on a crowd CNN model to estimate the crowd count of an image. Our method achieves the best MAE, and comparable MSE with existing methods. The reason is we divide the image into two parts, high-density people we use the dense map. It does not need to adjust too much parameters for the whole image. And for the foreground with low population density we use the detection system for counting to get a precise crowd number. This indicates that our model can estimate not only images with our dataset but also images with other crowd scenes. The result of this dataset is shown in Fig. \ref{fig:8}(b).

\subsection{The Mall dataset}
The Mall dataset was first introduced by Chen et al.~\citep{Chen2012Feature}. This dataset was captured using a publicly accessible surveillance camera in a shopping mall with challenging lighting conditions and glass surface reflections. And the dataset experiences more severe perspective distortion, which causes larger changes in size and appearance of objects at different depths of the scene, and has more frequent occlusion problems caused by the scene objects, e.g. stalls, indoor plants along the walking path. The Mall dataset also covers diverse crowd densities, from sparse to crowded, as well as different activity patterns(static and moving crowds) under a large range of illumination conditions at different times of the day.

By following the same setting with our method. This dataset does not satisfy assumptions that the crowd is evenly distributed. We choose the images that fit our scenes for evaluation. We delete pictures which fewer than 15 people in the mall dataset. Table 4 shows the results of our method and other methods on the mall dataset. We also achieve better MSE for this dataset. The dataset has fewer people but has a far-view region. This indicates that our model can estimate not only images with extremely dense crowds but also images with relatively sparse crowds. The result of this dataset is shown in Fig. \ref{fig:8}(c).

\begin{table}[h]
\centering
\renewcommand\arraystretch{1.5}
\caption{\upshape Comparing performances of different methods on Mall dataset}
\label{my-label}
\begin{tabular}{|p{4.25cm}|p{1.cm}|p{1.cm}|}
\cline{1-3}
Methods & MAE  & MSE      \\ \cline{1-3}
RR\citep{Chen2012Feature}      & 3.59 & 19.0     \\ \cline{1-3}
CA-RR\citep{Chen2013Cumulative}   & 3.43 & 17.7     \\ \cline{1-3}
GPR\citep{Chan2012Counting}    & 3.72 & 20.1     \\ \cline{1-3}
DigCrowd   & \textbf{3.21} & \textbf{16.4}     \\ \cline{1-3}
\end{tabular}
\end{table}

\section{Conclusion}
Crowd counting in public areas is one of the challenging issues in the study of crowd behavior and also a very important research subject in the field of public security. Since public areas often contain several different objects moving at the same time, this makes the sizes of these objects usually small with similar appearance on surveillance footage~\citep{Xu2017An}. Crowds move randomly causing serious occlusion of individuals. These factors make the analysis of crowd counting in public areas very difficult. In order to achieve efficient and accurate crowd counting, we propose acrowd estimation method (Depth Information Guided Crowd Counting (DigCrowd)) for the fusion of EDOF scenes. In order to demonstrate the performance of our system in an actual case of crowd counting, we produced a new airport dataset called CIISR. So far our model has shown excellent results on our dataset. And our model has good performance by fine-tuning the layers of the trained model, which shows that our approach can be applied to other dataset.


%

%
%
%
\section*{Acknowledgment}
This work is supported and funded by the National Natural
Science Foundation of China (Grant nos. 61602425, 61672469, 
61602420).We would like to thank the reviewers for their constructive comments
and suggestions.

\bibliographystyle{model2-names}
\bibliography{refs}
%
%

\end{document}